\documentclass[journal]{IEEEtran}
\ifCLASSINFOpdf
\else
\fi

\usepackage{epsfig}
\usepackage{graphicx}
\usepackage{amsmath}
\usepackage{amssymb}

\usepackage[sort, numbers]{natbib}
\usepackage{enumitem}
\usepackage{tablefootnote}
\usepackage{multirow}
\usepackage{capt-of}
\usepackage{fixltx2e}
\usepackage{floatrow}
\usepackage[pagebackref=true,breaklinks=true,letterpaper=true,colorlinks,bookmarks=false]{hyperref}
\newcommand{\x}{\times} 
\newcommand{\etc}{\textit{etc}} 
\newcommand{\ie}{\textit{i.e.,}} 
\newcommand{\etal}{\textit{et al.}} 
\newcommand{\R}{\mathbb{R}} 
\newcommand{\C}{\mathcal{C}} 
\newcommand{\sub}{\textsubscript} 
\newcommand{\UCF}{UCF\textunderscore CROWD\textunderscore 50 }

\begin{document}
%
\title{HA-CCN: Hierarchical Attention-based Crowd Counting Network}
%
%
%

\author{Vishwanath A. Sindagi,~\IEEEmembership{Student Member,~IEEE} and
        Vishal M. Patel,~\IEEEmembership{Senior Member,~IEEE}
         
\thanks{V. A. Sindagi and V. M. Patel are with the Department
of Electrical and Computer Engineering, Johns Hopkins University, Baltimore,
MD, 21218 USA e-mail: (vishwanathsindagi@jhu.edu, vpatel36@jhu.edul).}
}

%
%

\markboth{}%
{Shell \MakeLowercase{\textit{et al.}}: Bare Demo of IEEEtran.cls for IEEE Journals}
%



\maketitle

\begin{abstract}
Single image-based crowd counting has recently witnessed increased focus, but many leading methods are far from optimal, especially  in highly congested scenes. In this paper, we present Hierarchical Attention-based Crowd Counting Network (HA-CCN) that employs attention mechanisms at various levels to selectively enhance the features of the network. The proposed method, which is based on the VGG16 network, consists of a spatial attention module (SAM) and a set of global attention modules (GAM).  SAM enhances low-level features in the network by infusing spatial segmentation information, whereas the GAM  focuses on enhancing channel-wise information in the higher level layers. The proposed method is a single-step training framework,  simple to implement and achieves state-of-the-art results on different datasets.  

Furthermore, we extend the proposed counting network by  introducing a novel set-up to adapt the network to different scenes and datasets via weak supervision using image-level labels. This new set up reduces the burden of acquiring labour intensive point-wise annotations for new datasets while improving the cross-dataset performance. 



\end{abstract}


%
\IEEEpeerreviewmaketitle

\section{Introduction}
%
%
%
%
\IEEEPARstart{T}{he}  task of crowd counting is riddled with various challenges such as perspective distortion, extreme scale variations, heavy occlusion, illumination changes, clutter, non-uniform distribution of people, etc. Due to these issues, crowd counting and density estimation is a very difficult problem, especially in highly congested scenes. Several recent convolutional neural network (CNN) based methods for counting \cite{zhang2015cross,zhang2016single,sam2017switching,sindagi2017generating,liu2018leveraging,shi2018crowd_negative,shen2018adversarial,babu2018divide,cao2018scale,ranjan2018iterative,sam2018top,sam2019almost} have attempted to address one or more of these issues by adding more robustness to scale variations by proposing different techniques such as  multi-column networks \cite{zhang2016single}, intelligent selection of regressors suited for a particular crowd scenario \cite{sam2017switching} and incorporating global,  local context information into the counting network \cite{sindagi2017generating}, \etc. Methods such as \cite{sam2017switching,babu2018divide,sindagi2017generating} achieve significantly lower errors compared to the earlier approaches, however, they are complex to train due to the presence of multiple learning stages. For instance, Switching-CNN \cite{sam2017switching} involves different stages such as pre-training, differential training, switch training and coupled training. Similarly, CP-CNN \cite{sindagi2017generating} requires that their local and global estimators to be trained separately, followed by end-to-end training of their density map estimator.  Although the most recent methods such as \cite{ranjan2018iterative,shi2018crowd_negative,shen2018adversarial} achieve better results while being efficient, there is still considerable room for further improvements.

\begin{figure}[t!]
	\begin{center}
		\includegraphics[width=1\linewidth]{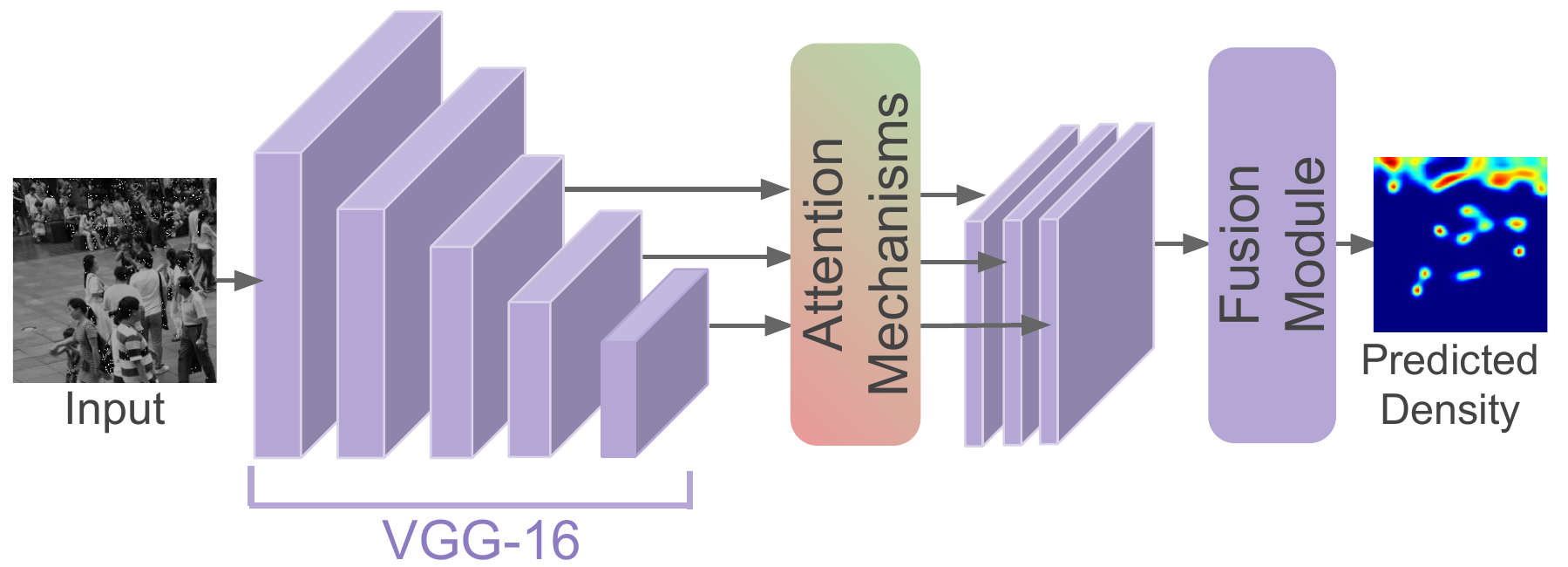}
	\end{center}
	\vskip -10pt \caption{Overview of the proposed Hierarchical Attention-based Crowd Counting Network (HA-CCN).}
	\label{fig:overview1}
 
\end{figure}

In this paper, we propose to improve the counting performance  by explicitly modeling spatial pixel-wise attention and global attention into the counting network. Considering that crowd images have large variations in head sizes, it is essential to leverage multi-scale information by employing feature maps from different conv layers of the VGG16 network \cite{simonyan2014very}. Several works such as \cite{lin2017feature,cai2016unified,sindagi2019dafe,sindagi2019mvx} have demonstrated that different sized objects are captured by different layers in a deep network.  Hence, an obvious approach would be to design a multi-scale counting network \cite{wang2018high,zhang2018multi} that concatenates feature maps from different layers of the VGG16 network. However, earlier layers in a deep network capture primitive features and do not learn semantic awareness. Due to this, naive concatenation of feature maps from different layers of the network is not necessarily an optimal approach to address the issue of large scale variations in crowd images.

To address this issue, we introduce a spatial attention module (SAM) in the network, that is designed to infuse semantic awareness into the feature maps. This module takes the  feature maps from lower layers as input, and  learns to perform foreground-background segmentation. Furthermore, it uses this learned segmentation map to enhance the lower layer feature maps by selectively attending to specific spatial locations in this lower layer.  Furthermore, we also attempt to augment channel-wise information in the higher level layers by employing a set of global attention modules. These modules  selectively enhance important channels while suppressing the unnecessary ones. Fig. \ref{fig:overview1} provides an overview of the proposed attention-based feature concatenation for multi-scale crowd counting.

\begin{figure}[t!]
	\begin{center}
		\includegraphics[width=1\linewidth]{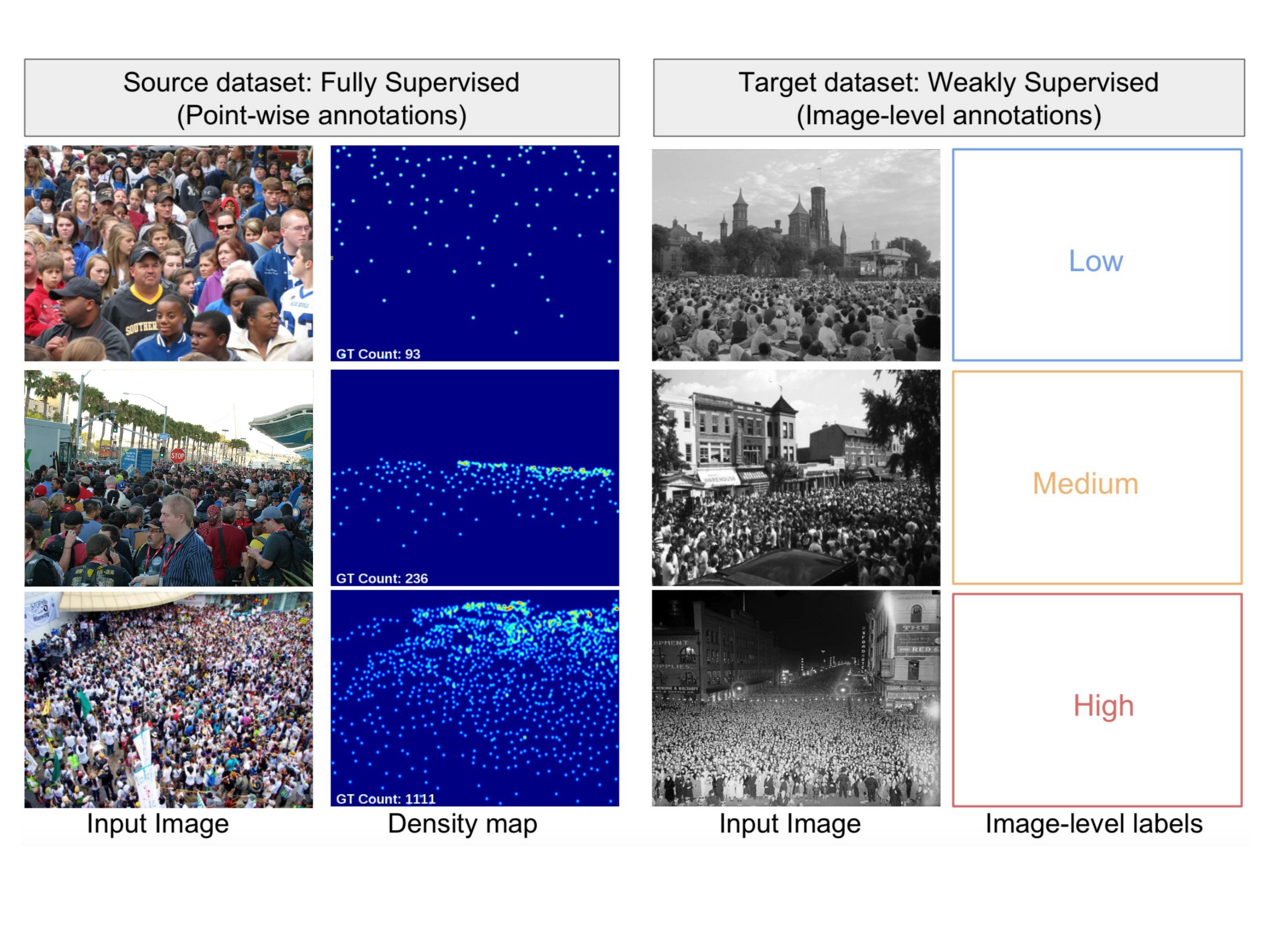}
		\vskip -25pt
		(a)\hskip 100pt(b)
	\end{center}
	\vskip -10pt \caption{Target dataset adaptation. (a) Source dataset with point-wise annotations is used to train the counting network. (b) Target dataset with only image-level annotations is used to fine-tune the pre-trained counting network.}
	\label{fig:overview2}
\end{figure}

In addition to improving the count performance, another major issue in the crowd counting research community is the poor generalization performance of the existing networks. This is due to the fact that CNN-based methods are highly data-driven and suffer from inherent dataset bias. Hence, they cannot be applied directly to new scenes without further fine-tuning. A simple solution to this would be to train the model on the target dataset in a fully-supervised fashion, which requires expensive ground-truth annotations.  Several earlier works such as \cite{zhang2015cross,liu2018leveraging} address this issue by proposing different semi-supervised or unsupervised fine-tuning methods in addition to their novel network designs. For instance, Zhang \etal \cite{zhang2015cross} presented a cross-scene counting approach where they use perspective maps to retrieve candidate scenes from source dataset that are similar to the target set, which are then used to fine-tune the network. However, perspective maps may not be always available. Additionally, it is dependent on the assumption that their pre-trained model provides good estimates of count in the target patches. Liu \etal \cite{liu2018leveraging} proposed a self-supervised method based on image ranking to adapt to different datasets. While it achieves better generalization performance, their method is still limited since they use only unlabeled data. 

To address this generalization issue, we take a different approach as compared to earlier attempts (\cite{zhang2015cross,liu2018leveraging}) by proposing  a novel weakly supervised learning setup. We leverage image-level labels, which are much easier to obtain as compared to point-wise annotations \footnote{Crowd counting datasets are usually provided with point-wise (x,y) location annotations, which are converted to pixel-wise density maps.},  in a weakly supervised fashion for fine-tuning networks to newer datasets/scenes. To achieve this weak supervision, we use the idea of image-level labeling of crowd images into different density levels by Sindagi \etal \cite{sindagi2017generating} and Fu \etal \cite{ fu2015fast}. While these methods \cite{sindagi2017generating, fu2015fast} employ image-level labels in conjunction to point-wise annotations to train their networks, we propose to use only image-level labels in the weakly supervised setup while adapting to new datasets, thereby avoiding the labour intensive point-wise annotation process.  Fig. \ref{fig:overview2} illustrates the different types of annotations used for training the network. Fig. \ref{fig:overview2}(a) represents samples from a source dataset, which consists of images and corresponding point-wise ground-truth annotations. The source dataset is used to pre-train the counting network. Fig. \ref{fig:overview2}(b) represents samples from the target set to which we intend to adapt the pre-trained counting network. The pre-trained network is then fine-tuned on the target dataset using image-level labels via the proposed weakly supervised approach. During testing, the estimated density map from the fine-tuned network is compared with the the ground-truth using standard metrics as described later. 

To summarize, the following are our key contributions in this paper:
\begin{itemize}[topsep=0pt,noitemsep,leftmargin=*]
	\item A new network design that employs  attention mechanisms at various levels for selectively enhancing the features from different layers of VGG16 network to increase the effectiveness of multi-scale concatenation. 
	\item  A novel setup to adapt existing crowd counting models to new scenes and datasets via weakly supervised learning. To the best of our knowledge, this is the first attempt to perform weak supervision using image-level labels for crowd counting. 
\end{itemize}

In the following sections, we discuss related work (Section \ref{sec:related}) and the proposed method in detail (Section \ref{sec:method}). Details of experiments and results of the proposed method along with comparison on different datasets are provided in Section \ref{sec:experiments}, followed by conclusions in Section \ref{sec:conclusion}.

\section{Related work}
\label{sec:related}

\noindent\textbf{Crowd Counting.} Early approaches for crowd counting are based on hand-crafted representations and  different regression techniques \cite{li2008estimating,ryan2009crowd,chen2012feature,idrees2013multi,lempitsky2010learning,pham2015count,xu2016crowd}.  A  comprehensive survey of these early methods  can be found in \cite{chen2012feature,loy2013crowd,li2015crowded}. Recent focus in the crowd counting community has been towards exploiting the advances in CNN-based methods and in this attempt, methods such as \cite{wang2015deep,zhang2015cross,sam2017switching,arteta2016counting,walach2016learning,onoro2016towards,zhang2016single,sam2017switching, sindagi2017cnnbased, sindagi2017generating,boominathan2016crowdnet, onoro2018learning,wang2018indefense,sindagi2017survey,liu2018decidenet,wang2018defense,onoro2018learning} have demonstrated significant improvements over the traditional methods. Majority of the existing work is focused on addressing the problem of large scale variations in crowd images through different techniques such as multi-resolution network \cite{onoro2016towards}, multi-column networks \cite{zhang2016single}, selective regression \cite{sam2017switching}, context-aware counting \cite{sindagi2017survey}.

Babu \etal \cite{babu2018divide} proposed an automatically growing CNN to progressively increase the capacity of the network based on the dataset. Shen \etal \cite{shen2018adversarial} used adversarial loss similar to \cite{sindagi2017generating} to attenuate blurry effects in the estimated density maps.  Shi \etal \cite{shi2018crowd_negative} proposed deep negative correlation based learning of more generalizable features. In another interesting approach, Liu \etal \cite{liu2018leveraging} proposed to leverage unlabeled data for counting by introducing a learning to rank framework. 
Li \etal \cite{li2018csrnet} proposed CSR-Net, that consists   of two  components: a front end CNN-based feature extractor and  a dilated CNN for the back-end. Ranjan \etal \cite{ranjan2018iterative} proposed a network with two branches that estimates density map in a cascaded fashion.  Cao \etal \cite{cao2018scale} proposed an encoder-decoder network with scale aggregation modules. They use a combination of Euclidean loss and a newly introduce local pattern consistency loss to train their network. Idrees \etal \cite{idrees2018composition} proposed a new large-scale crowd dataset with 1.25 million annotations, along with a novel loss function to train their dense-net \cite{huang2017densely} based architecture. 

Recently, a few approaches have been proposed that incorporate detection of crowded regions through different techniques such as attention injective deformable network \cite{liu2018adcrowdnet}, semantic prior-based residual regression \cite{wan2019residual}, use of auxiliary task such as segmentation \cite{zhao2019leveraging} and segmentation infusion via inverse attention \cite{sindagi2019inverse}. Other techniques such as \cite{jiang2019crowd} exploit multi-scale features in different ways. For instance, Jiang \etal \cite{jiang2019crowd,shi2019revisiting,liu2019context} propose a Trellis style encoder decoder, where the multi-scale feature maps in the decoder are combined in an effective way. Similarly, Shi \etal and Liu \etal \cite{shi2019revisiting,liu2019context} exploit multi-scale features by explicitly considering perspective and context information respectively.  Zhang \etal \cite{zhang2019wide} address the issue of wide area counting by proposing a multi-view fusion CNN. Wang \etal \cite{wang2019learning} presented a new  large-scale, diverse synthetic dataset and proposed a SSIM based CycleGAN \cite{zhu2017unpaired} to adapt the synthetic datasets to real world dataset. \\

\begin{figure*}[ht!]
	\begin{center}
		\includegraphics[width=.91\linewidth]{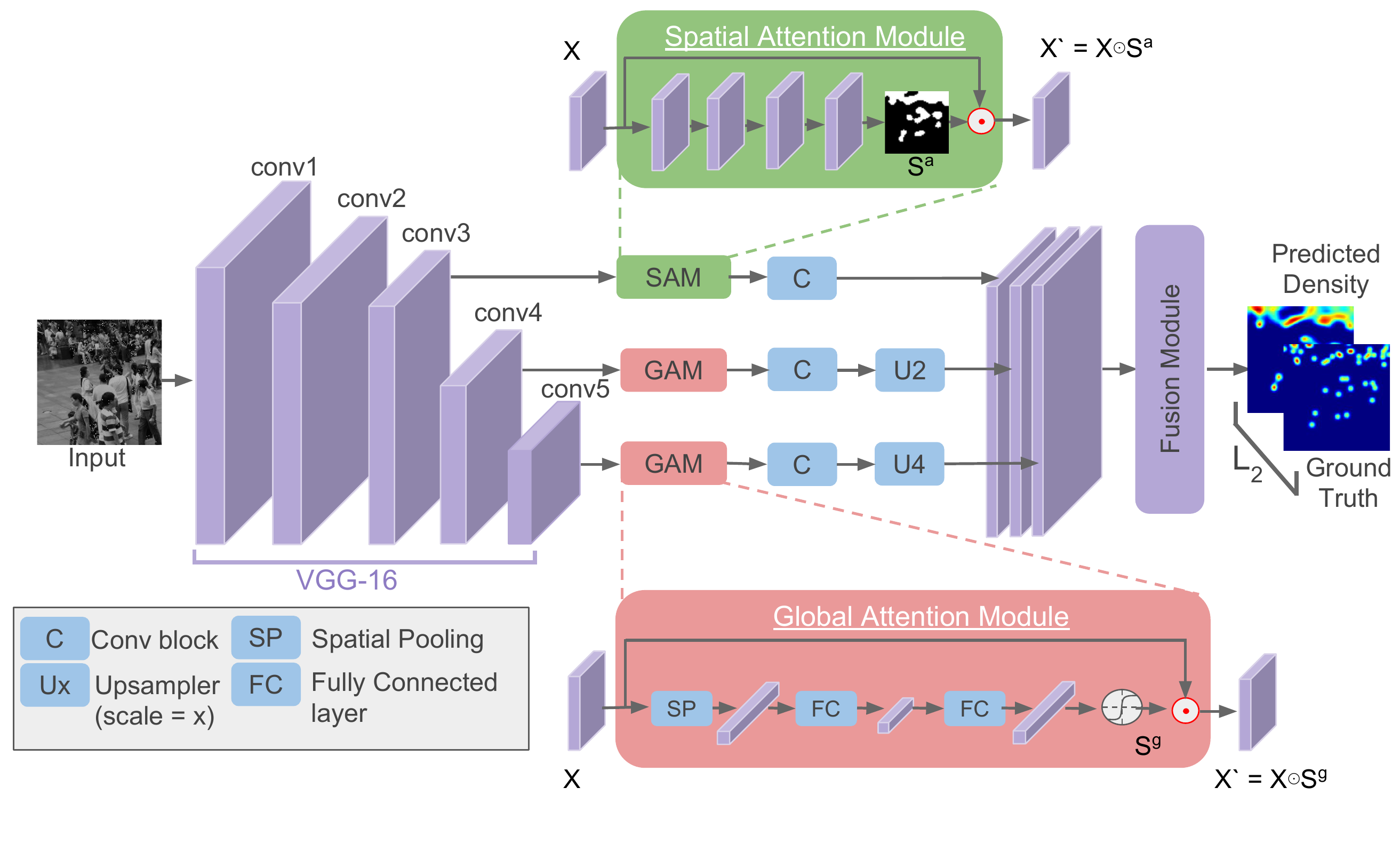}
	\end{center}
	\vskip -35pt \caption{Overview of the proposed Hierarchical Attention-based Crowd Counting Network (HA-CCN). VGG16 is used as the base network. Feature maps from conv3 are forwarded through a spatial attention module that incorporates pixel-wise segmentation information into the features. Feature maps form higher layers (conv4, conv5) are forwarded through a set of global attention modules that augment the feature maps along the channel dimension.}
	\label{fig:arch}
\end{figure*}

\noindent\textbf{Attention mechanisms.} Inspired by the role of attention in human visual perception \cite{itti1998model,corbetta2002control,tang2015neuroscience}, several works have successfully incorporated  attention mechanism to improve the performance of CNNs for a variety of tasks such as   image captioning \cite{wang2018bidirectional,chen2017sca}, visual question answering \cite{yang2016stacked,chen2015abc,anderson2018bottom,xu2016ask},  pose-estimation \cite{chu2017multi}, classification \cite{woo2018cbam,hu2018gather,park2018bam}, detection \cite{zhang2018progressive}, fine-grained recognition \cite{zheng2017learning}, sequence to sequence modeling \cite{zhu2017bidirectional} \etc. Xu \etal \cite{xu2016ask} were among the first to introduce visual attention in image captioning where they use different pooling mechanisms that attend to important and relevant regions in the scene.   Zhu et al. \cite{zhu2016visual7w} employed  soft attention to combine image region features for the task of VQA. For the same task, Yang et al. \cite{yang2016stacked} and Xu et al. \cite{xu2016ask} employed multiple stacked spatial attention models, in which the spatial attention map is successively refined. Chu \etal. \cite{chu2017multi} used attention to guide multi-contextual representation learning for improving pose estimation performance.  Wang et al. \cite{wang2017residual} improved classification performance by incorporating 3D attention maps, generated using hour glass modules, into  residual  networks. Hu et al. \cite{hu2017squeeze}  proposed a compact Squeeze-and-Excitation (SE) module to leverage  inter-channel relationships.  Recently, Woo \etal \cite{woo2018cbam} extended the work of \cite{park2018bam} by employing spatial and channel-wise attention modules after every layer in the network.  In a different application, Zhu \textit{et al.} \cite{zhu2017bidirectional} employed seq2seq in their decoder structure to model temporal video sequences. The attention mechanism at each step is used to help the decoder to decide which frames in the input sequence might be related to the next frame reconstruction. Note that they use attention mechanism to perform input selection. In contrast, we use attention mechanism to select relevant and important features and additionally refine them.

The closest methods to our work are \cite{woo2018cbam,chen2017sca,zhu2017bidirectional}. There are several notable differences as compared to our method. First, these methods employ a sequence of channel-wise and spatial attention after every convolutional layer to refine the feature maps. In contrast, we specifically insert a spatial attention module after the conv3 block. By doing this,  we are able to infuse the attention early into forward process and hence, such a module is rendered unnecessary after every block for an application like crowd counting. Second, the spatial attention modules in the existing works are self-supervised. Different from them, we employ a mask guided spatial attention module that is explicitly supervised   using foreground/background masks, resulting in faster learning of the spatial attention. Further, the global attention modules are inserted only after conv4 and conv5 blocks. These different modules are carefully added at specific blocks in the network, thereby avoiding unnecessary over-parameterization.\\

\noindent\textbf{Weak Supervision.} Weakly supervised learning has been extensively used for various problems in computer vision such as semantic segmentation \cite{saleh2016built,chaudhry2017discovering,pinheiro2015image,roy2017combining,hong2017weakly,papandreou2015weakly}, object localization \cite{cinbis2017weakly,oquab2015object,kantorov2016contextlocnet,shi2017weakly}, saliency detection \cite{zhang2017supervision,wang2017learning}, scene recognition \cite{oquab2014weakly,wang2017weakly} and many more. However, this form of learning has been relatively unexplored for crowd counting. Liu \etal \cite{liu2015bayesian} proposed a solution based on Bayesian model adaptation of Gaussian processes for transfer learning, which is limited only to the GP model of \cite{chan2012counting}. Recently, Borstel \etal \cite{von2016gaussian} proposed a method to count objects within an image from only region-level count information. Though the problem is defined in a weakly supervised setting, they require local region level count as annotations which is a labour intensive process.

In this work, we introduce a novel weakly-supervised learning setup  that employs  image-level labels to generate pseudo ground-truth, which is further used to fine-tune the counting network.  The use of pseudo ground-truth generation is inspired from semi-supervised learning approaches such as \cite{wu2019progressive,liu2017provid,kingma2014semi,rasmus2015semi,dong2018few,kipf2016semi}. These approaches typically use labeled data in the dataset to train a predictor. The trained predictor is then used to generate predictions  for the unlabeled data. The highly confident predictions are then used as ground-truth to further fine-tune the data.  In contrast to this framework, we propose a weakly supervised approach where the pseudo-ground truth is generated from weak image-level labels.

\section{Hierarchical attention for  crowd counting}
\label{sec:method}

As discussed earlier, a natural solution to address scale variation in crowd counting images is to leverage multi-scale features from different layers in the backbone network. However, the layers in the backbone network  are learned in a hierarchical manner with the earlier layers capturing primitive features and the subsequent layers capturing higher level concepts. Hence, direct fusion of these multi-scale feature maps might not be the most effective approach. In order to overcome this, we propose Hierarchical Attention-based Crowd Counting Network  (HA-CCN)  that leverages attention mechanisms to enrich features from different layers of the network for more effective multi-scale fusion.  

Fig. \ref{fig:arch} provides an overview of the proposed method, which is based on the VGG-16 network. We include a spatial attention module (SAM) and a set of global attention modules (GAM)  with the objective of enriching the feature maps at different levels. The base network consists of conv layers (conv1 $\sim$ conv5) from the VGG16 network. The conv3 features are enhanced by passing them through SAM. Similarly, features from conv4 and conv5 are passed through GAMs in order to perform channel-wise enhancement. The enhanced feature maps from conv3 are then forwarded through a conv block which consists of 3 conv layers defined as follows:
\{Conv2d(256,64,1)$^{\ref{fn:conv}}$-ReLU, 
Conv2d(64,64,3)$^{\ref{fn:conv}}$-ReLU, 
Conv2d(64,24,1)$^{\ref{fn:conv}}$-ReLU\}.

Similarly, the enhanced features from conv4 and conv5 are forwarded through a conv block  and an upsampling layer to scale the feature maps to a size similar to that of conv3 feature maps. The conv block is defined by:
\{Conv2d(512,64,1)$^{\ref{fn:conv}}$-ReLU, 
Conv2d(64,64,3)$^{\ref{fn:conv}}$-ReLU, 
Conv2d(64,24,1)$^{\ref{fn:conv}}$-ReLU\}.

These processed features are  concatenated together before being forwarded through the fusion module that consists of a set of conv layers to produce the final density map. These conv layers are  defined by:
\{Conv2d(72,64,1)$^{\ref{fn:conv}}$-ReLU, 
Conv2d(64,64,3)$^{\ref{fn:conv}}$-ReLU, 
Conv2d(64,1,1)$^{\ref{fn:conv}}$-ReLU\}.
The network is trained by minimizing the Euclidean distance between the predicted density map and the ground truth density map as below:
\begin{equation}
\label{eq:lossdensity}
L_d = \frac{1}{N}\sum_{i=1}^{N}\|F_d(X_i,\Theta) - D_{i}\|_2,
\end{equation}
where, $N$ is number of training samples, $X_i$ is the $i$\textsuperscript{th} input image, $F_d(X_i, \Theta)$ is the estimated density,  $D_i$ is the $i$\textsuperscript{th} ground-truth density and it is calculated by summing a 2D Gaussian kernel centered at every person's location $x_g$ as follows: $D_i(x) = \sum_{{x_g \in S}}\mathcal{N}(x-x_g,\sigma)$,
where $\sigma$ is scale parameter of 2D Gaussian kernel and $S$ is the set of all points where people are located.  The density map generated by the network is $1$/$4^{th}$ of the input image resolution. Due to its construction, the sum of the density map provides  an estimate of the number of people in the input image.

Details of the proposed method and its various components are described in the following sub-sections.

\begin{figure*}[ht!]	
	\centering	
	\includegraphics[width=5.6cm]{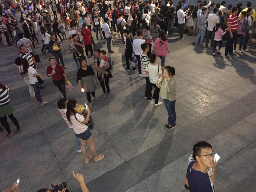}
	\includegraphics[width=5.6cm]{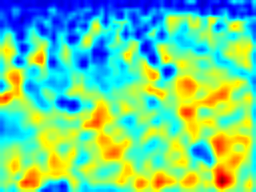}
	\includegraphics[width=5.6cm]{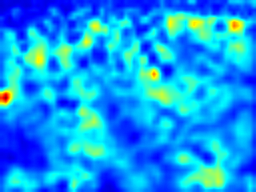}\\
	\vspace{5pt}
	\includegraphics[width=5.6cm]{figures/fbefore/inoutput_32.png}
	\includegraphics[width=5.6cm]{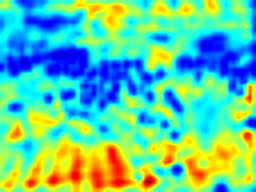}
	\includegraphics[width=5.6cm]{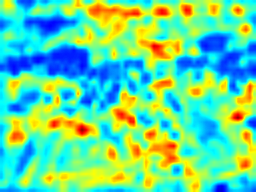}
	
	(a) \hskip150pt(b) \hskip150pt(c) 
	\vskip -10pt\caption{Visualization of the conv3 feature maps: (a) Input image (b) Before segmentation infusion (c) After segmentation infusion. By infusing segmentation information into the counting network, we are able to suppress background regions. Note that in the density maps,  red color indicates high density and blue color indicates low density. }
	\label{fig:fmapmany}	
\end{figure*}

\subsection{Spatial attention module}
\label{ssec:segmentation}

Inspired by the success of spatial attention mechanisms in image captioning, visual question answering and classification \cite{zheng2017learning,xu2016ask,chen2017sca}, we explore its utilization for crowd counting. The goal of spatial attention is to select attentive regions in the feature maps, which are then used to dynamically enhance the feature responses. In contrast to  existing work that  learn   spatial attention in a self-supervised manner, we propose to learn this by explicitly using  foreground background segmentation for supervision.  Since the goal of the spatial attention module is to focus on relevant regions and foreground regions are necessarily a part of these relevant regions, it is beneficial to use these labels to supervise the module. By explicitly supervising the module, we are able to infuse foreground background information into the network, thereby forcing the network to focus on relevant regions among the foreground. Moreover, these labels are readily available and hence, it does not require additional annotation efforts.

The spatial attention module consists of 4 conv layers with 3$\x$3 filters that takes feature maps  from the conv3 layer of the VGG16 network as input (denoted by $X \in \R^{W\x H\x C}$), and produces a segmentation output $S ^a\in [0,1]^{W\x H}$.  The segmentation map is then used to actuate the low  level feature map $X$ via element-wise multiplication: $\hat{X} = X\odot S^a$, where $\hat{X}$ is the actuated low level feature map from conv3. Through this attention mechanism, we are able to incorporate segmentation awareness into the low level feature maps. As illustrated in Fig. \ref{fig:fmapmany}, the use of segmentation information into the network enriches feature maps by suppressing irrelevant regions and boosting the foreground regions. The actuated feature maps are then forwarded to the fusion block (FM), where they are fused with the features from other layers to generate the final density map.

The weights of SAM are learned by minimizing the cross entropy error between the predicted segmentation map and the corresponding ground-truth. Normally, the segmentation task requires pixel-wise annotations. However, in this case  existing ground truth density map annotations are thresholded to generate the ground truth segmentation maps, which are then used to train the spatial attention module. Basically, the pixels that contain head regions are labeled as 1 (foreground), and otherwise as 0 (background).  Hence, the proposed method  does not require any additional labeling. In spite of these annotations being noisy, the use of segmentation information results in considerable gains.

\begin{figure*}[h!]
	\begin{center}
		\includegraphics[width=1\linewidth]{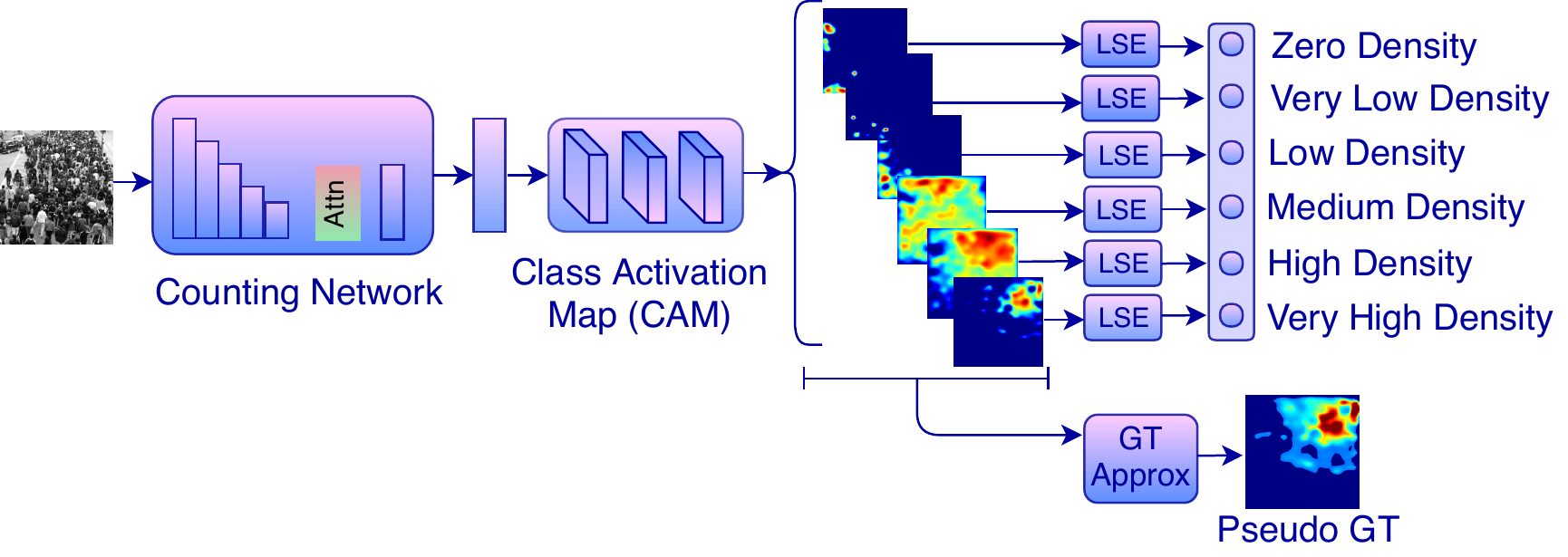}
	\end{center}
	\vskip -15pt \caption{Overview of the proposed weakly supervised learning for adapting counting network to new datasets. A class activation map (CAM) module  is learned to produce class-wise score maps using image-level labels, which are further used to estimate pseudo ground-truth density maps for target set images.}
	\label{fig:wsl}
\end{figure*}

\subsection{Global attention modules}
\label{ssec:hsm}

In contrast to  the spatial attention module that attends to relevant spatial locations in the feature maps of low-level layers, the global attention module (GAM) is designed to attend to feature maps in the channel-dimension. The global attention module is similar to the channel-wise attention used in earlier work like \cite{chen2017sca,woo2018cbam}. Specifically, this module consumes feature maps from the backbone network and learns to compute attention along the channel dimension. The computed attention captures the important channels in the feature maps and hence aids in suppressing   information from unnecessary channels. Since this module operates at a global level in terms of spatial dimension, we refer to this attention module as global attention module. It has been demonstrated in \cite{chen2017sca,zheng2017learning,hu2017squeeze}, that channels capture the presence either different parts of an object or different classes of objects and channel-wise attention is an effective way to boost the correlation between object/object parts and image captions. 

Based on these considerations, we employ a set of global attention modules, which take feature maps from the higher conv layers as input and produce a channel-wise attention map, which is then used to actuate the feature maps along the channel dimension. Mathematically, given a feature map input $X \in \R^{W\x H\x C}$, GAM first performs a spatial pooling to produce pooled features $Y \in \R^{1\x 1\x C}$ using 
\begin{equation}
Y_i  =\frac{1}{W\x H} \sum_{w,h}X_{i}^{wh},
\end{equation}

where $i$ is the channel index, and $w,h$ are spatial indices. 
$Y$ is passed through a set of fully-connected (FC) layers defined by $FC(512,64)-ReLU-FC(64,64)-ReLU-FC(64,512)$\footnote{ \label{fn:fc}\textit{FC\sub{N\sub{i},N\sub{o}}} denotes fully connected  layer (with \textit{N\sub{i}} input elements, \textit{N\sub{o}} output elements)} and  a sigmoid layer to produce channel-wise attention vector $S^g\in \R^{1\x 1\x C}$.  Finally, $S^g$ is  used to actuate the feature maps from the higher layer by performing a element-wise multiplication, \ie $\hat{X} = X\odot S^g$.

\section{Weak supervision via image-level labels}
\label{sec:wsl}

As discussed earlier, existing methods \cite{zhang2016single,zhang2015cross} recognize the inability of these networks to generalize well to different datasets. Their solutions to improve the cross-dataset performance is through fine-tuning in either a fully-supervised or semi-supervised fashion. In contrast to these approaches, we propose a weakly supervised setup to train the counting networks on the new datasets with just image-level labels. Such a setup will simplify the training process as it does not require point-wise annotations which are labour intensive and expensive to obtain. 

The idea of performing weakly supervised crowd counting is largely inspired by the success of recent CNN-based weakly supervised semantic segmentation methods \cite{pinheiro2015image,saleh2016built,chaudhry2017discovering} that typically fit the problem into Multiple-Instance Learning framework \cite{maron1998framework}. In their setup, every image is considered to have at least one pixel corresponding to image class label, and the segmentation task is formulated as inferring the pixels belonging to the object class. These methods usually employ class activation mappings to perform weak supervision. However, crowd counting is a regression problem and cannot be directly fit into such a framework. To overcome this issue, crowd counting is transformed into a crowd-density classification task, i.e., instead of counting the number of people in an image, this task is reformulated into categorizing the image into one of the six classes: $\C$ =\textit{ \{zero density, very-low density, low density, medium density, high density, very-high density\}}. This reformulation is based on the intuition that it is easier to label an image as containing large or few number of people as compared to the exact count. Sindagi \etal \cite{sindagi2017generating} used a similar concept for leveraging image context. In this work, the labels are used to reformulate the counting problem into a classification task for weakly supervised learning.

\begin{figure*}[ht!]
	\begin{center}
		\includegraphics[width=.91\linewidth]{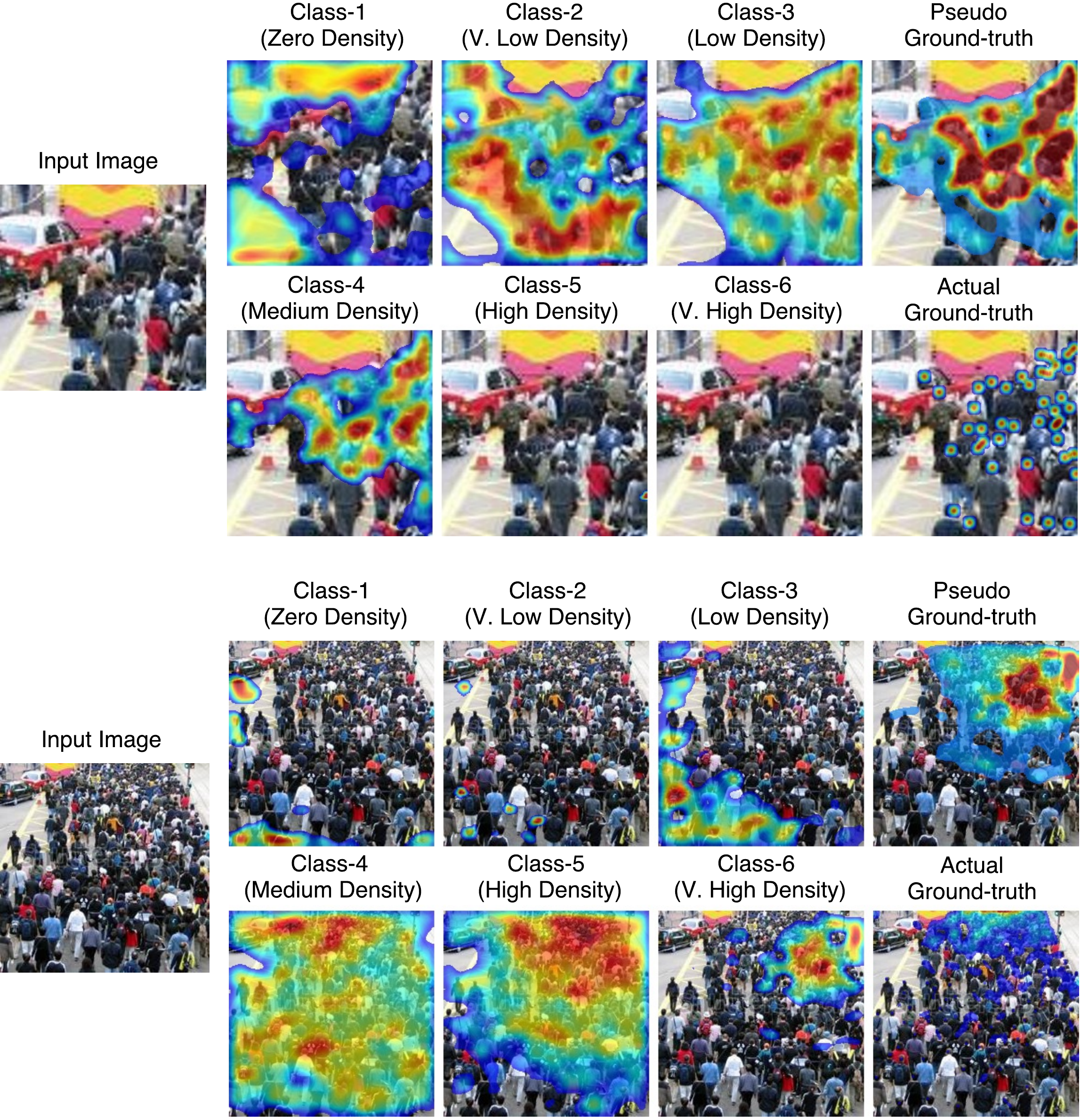}
	\end{center}
	\vskip -10pt \caption{Example of class-wise score maps overlaid on input images. It can be observed that the CAM module is able to accurately identify regions corresponding to different density levels in an image. We also illustrate pseudo ground-truth estimated using image-level labels. Note that in the density maps,  red color indicates high density and blue color indicates low density. }
	\label{fig:segmaps}
\end{figure*}

Fig. \ref{fig:wsl} illustrates the proposed weakly supervised approach for adapting to new target scenes or datasets. Similar to semantic segmentation where a pre-trained CNN is used, we use counting network (HA-CCN) described in Section \ref{sec:method} that is pre-trained on the source dataset.  A class activation map module (CAM), consisting of 4 conv layers   is added before the fusion module in the counting network. This module is defined as:
\{Conv2d(72,64,3)-ReLU, 
Conv2d(64,64,3)-ReLU, 
Conv2d(64,32,3)-ReLU\}.
Conv2d(32,6,3)\footnote{ \label{fn:conv}\textit{Conv2d(N\sub{i},N\sub{o},k)} denotes 2d convolutional layer (with \textit{N\sub{i}} input channels, \textit{N\sub{o}} output channels, \textit{k}$\times$\textit{k} filter size)}

This sub-network takes in features from the counting network and processes them to produce output with $|\C|$ feature planes, one for each class. That is, the output of CAM is pixel-wise scores for each class and is denoted by $S^c_{i,j}$ at pixel location $(i,j)$ for each class $c\, \epsilon \, \C$.  Since point-wise labels are not available for the target set, the pixel-wise scores for each class are mapped to a single image-level classification score using an aggregation function $F_{agg}$ such that $s^c = F_{agg}(S^c_{i,j})$. This class-wise ($s^c$) score is then maximized for the right class label. Different aggregation functions such as Global Average Pooling (GAP) and Global Max Pooling (GMP) \cite{zhou2016learning} have been used in the literature. In case of GAP, all pixels in the score map are assigned the same weights even if they do not belong to image's class label. GMP addresses this by assigning weight to the pixel that contributes most to the score, however the training is slow \cite{pinheiro2015image}. Hence, smooth version and convex approximation of the max function is chosen for $F_{agg}$, called Log-Sum-Exp (LSE) which is defined as: 
\begin{equation}
\label{eq:agg}
S^c = \frac{1}{r}\log \left[\frac{1}{wh}\sum_{i,j}\left(rS^c_{i,j}\right)\right],
\end{equation}
where, $S^c$ denotes aggregated score for class $c$, $S^c_{i,j}$ is pixel-level score at location $(i,j)$ for class $c$, $r$ is a hyper-parameter that controls the smoothness of approximation, $w$,\,$h$ are width and height of the score map. A soft-max function is applied to the aggregated class scores. The CAM module is trained using the standard binary cross entropy loss function. Parameters of the counting network are kept fixed during this training. The class-wise score maps obtained from the above procedure indicate regions/pixels in the image that belong to a particular density level and hence can be viewed similar to class activation maps \cite{zhou2015learning} (see Fig. \ref{fig:segmaps}). These class-wise maps are then used to approximate the pseudo ground-truth density maps for the target set using: 
\begin{equation}
\label{eq:gtapprox}
D_{pseudo}(i,j)  = \sum_{c\epsilon \C}n(c)\widetilde{S}^c_{i,j},
\end{equation}
where, $\widetilde{S}^c_{i,j}$ are obtained by normalizing $S^c_{i,j}$ and $n(c)$ is the average count for class $c$ in the source dataset. The pseudo ground-truth maps (as seen in Fig. \ref{fig:segmaps}) are not as sharp as actual ground-truth maps, however, they provide coarse regional density that is better as compared to just image-level labels.

These pseudo ground-truth density maps are used to supervise the counting network on the target dataset. During fine-tuning, weights of the VGG-16 network are fixed and only the  weights of the later conv layers are updated. This ensures that the resulting estimated density maps are sharper since the feature maps extracted from VGG-16 preserve details, while the later layers adapt to the newer dataset. 

Although the network is trained using image-level labels, it learns to generate density maps for the target set as well. Hence, during inference, test image from the target set is forwarded through the network to estimate the density map. The performance of the proposed weakly supervised technique is measured  using standard count error metrics (MAE/MSE). 
\section{Experiments and results}
\label{sec:experiments}

\subsection{Hierarchical attention-based counting}

In this section, we first describe the training and implementation specifics followed by a detailed ablation study to understand the effects of different components in the proposed counting network.  Finally, we compare results of the proposed method against several recent approaches on 3 publicly available datasets (ShanghaiTech \cite{zhang2016single}, UCF-QNRF \cite{idrees2018composition},  \UCF \cite{idrees2013multi}).\\

\subsubsection{Training and implementation details}

\noindent The network is trained end-to-end using the Adam optimizer with a learning rate of 0.00005 and a momentum of 0.9 on a single NVIDIA GPU Titan Xp. 10 \% of the training set is set aside for validation purpose. The final training dataset is formed by cropping patches of size 224$\times$224 from 9 random locations from each image. Further data augmentation is performed by randomly flipping the images (horizontally) and adding random noise. Similar to earlier work, the count performance is measured using mean absolute error ($MAE$) and mean squared error ($MSE$) given by: 

\begin{equation}
\label{eq:count_error}
\nonumber  MAE = \frac{1}{N}\sum_{i=1}^{N}|y_i-y'_i|, \;\;
MSE = \sqrt{\frac{1}{N}\sum_{i=1}^{N}|y_i-y'_i|^2},
\end{equation}

where  $N$ is the number of test samples, $y_i$ is the ground-truth count and $y'_i$ is the estimated count corresponding to the $i^{th}$ sample.

Since the network is fully convolutional, entire test image is forwarded through the network during inference. This results in faster inference as compared to the existing methods (such as Switching-CNN \cite{sam2017switching}, IG-CNN \cite{babu2018divide}, CP-CNN \cite{sindagi2017generating}, SA-Net \cite{cao2018scale}) which involve patch-based testing.\\ 


\subsubsection{Architecture ablation}
\label{sssec:ablation}
\noindent To understand the effectiveness of  various modules present in the network, we perform experiments with the  different settings using ShanghaiTech dataset (Part A and Part B). This dataset consists of 2 parts with Part A containing 482 images and Part B containing 716 images and a total of 330,165 head annotations. Both parts have training and test subsets.

\begin{figure}[t!]
	\begin{center}
		\includegraphics[width=1\linewidth]{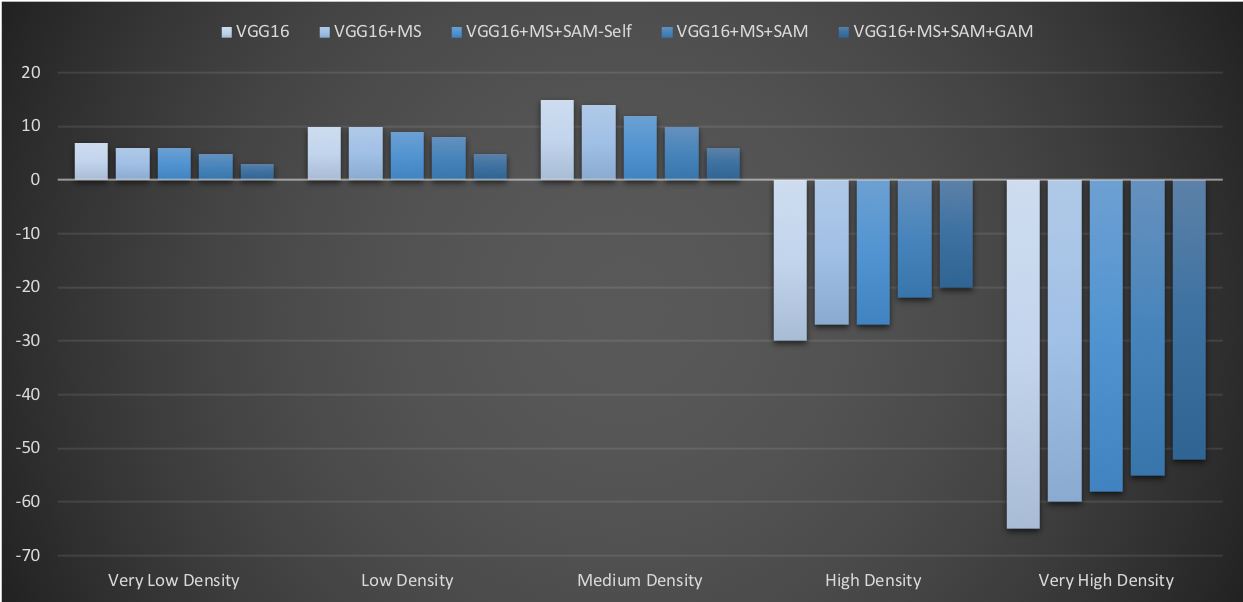}
		\vskip -20pt
	\end{center}
	\vskip -10pt \caption{Ablation study: MAE for different configurations at different density levels.}
	\label{fig:chart}
\end{figure}

\begin{table}[t!]
	\centering
	\caption{Results of the ablation study on ShanghaiTech Part A and Part B datasets.}
	\label{tab:ablation}
	\resizebox{1\linewidth}{!}{
		\begin{tabular}{|l|c|c|c|c|}
			\hline
			& \multicolumn{2}{|c|}{Part A}  & \multicolumn{2}{|c|}{Part B}  \\ \hline
			Configuration          & MAE          & MSE           & MAE          & MSE           \\\hline
			VGG16           & 78.3         & 120.1         & 18.3         & 22.9          \\
			VGG16+MS      & 72.1  & 115.5 & 15.6  & 20.6 \\
			VGG16+MS+SAM  (Self-sup)     & 69.5 & 108.2 & 12.3  & 20.1  \\
			VGG16+MS+SAM      & 65.1 & 103.5 & 10.6  & 19.6  \\
			VGG16+MS+SAM+GAM (HA-CCN) & \bf{62.9}  & \bf{94.9} & \bf{8.1}  & \bf{13.4} \\ \hline
		\end{tabular}
	}
\end{table}

The ablation study consisted of evaluating 3 baselines in addition to the proposed method:\\ (i)\textit{VGG16}: VGG16 network with an additional conv block at the end.\\
(ii)\textit{VGG16+MS}: VGG16 with multi-scale feature map concatenation and a fusion module at the end.\\
(iIi)\textit{VGG16+MS+SAM (Self-sup)}: VGG16 with spatial attention module (self-supervised) for conv3 layer and multi-scale feature map concatenation, followed by a fusion module at the end.\\
(iv)\textit{VGG16+MS+SAM}: VGG16 with spatial attention module for conv3 layer and multi-scale feature map concatenation, followed by a fusion module at the end.\\
(v)\textit{VGG16+MS+SAM+GAM (HA-CCN)}: proposed method. \\

The results of these experiments are tabulated in Table \ref{tab:ablation}. It can be observed that the naive approach of performing multi-scale feature concatenation does not necessarily yield the most optimal performance. The use of SAM infuses segmentation information in to the feature maps of conv3 layer in the base network, resulting in considerable reduction of the count error as compared to the naive approach. The use of global attention results in further improvement, thus showing significance of incorporating channel-wise importance in the network. 

Additionally, it can also be noted that the explicitly supervised SAM results in better performance as compared to the self-supervised spatial attention.

Fig. \ref{fig:chart} shows a plot of  the mean absolute error for different configurations in the ablation study at different density levels. It can be observed that the proposed HA-CCN network achieves best error among all the density levels.\\

\begin{table}[t!] 
	\centering 		
	\caption{Comparison of results on the ShanghaiTech \cite{zhang2016single} and \UCF \cite{idrees2015detecting} datasets. Top two methods are highlighted using underline and bold fonts respectively. * indicates patch-based testing.}
	\label{tab:shtech}
	\resizebox{1\linewidth}{!}{
		\begin{tabular}{|l|c|c|c|c|c|c|}
			\hline
			& \multicolumn{2}{c|}{ShTech-A} & \multicolumn{2}{|c|}{ShTech-B} & \multicolumn{2}{|c|}{UCF-CROWD}\\ \hline
			Method          & MAE          & MSE          & MAE          & MSE  & MAE          & MSE         \\ \hline
			Switch-CNN \cite{sam2017switching}* (CVPR '17)          & 90.4        & 135.0        & 21.6         & 33.4  & 318.1         & 439.2        \\ 
			CP-CNN \cite{sindagi2017generating}* (ICCV '17)          & 73.6        & 106.4        & 20.1         & 30.1  & 295.8         & \underline{\bf{320.9 }}       \\ 
			IG-CNN \cite{babu2018divide}* (CVPR '18)         & 72.5        & 118.2        & 13.6         & 21.1 & 291.4         & 349.4        \\ 
			ACSCP \cite{shen2018adversarial} (CVPR '18) & 75.7 & {{102.7}}  & 17.2 & 27.4 & 291.0 & 404.6 \\
			CSRNet \cite{li2018csrnet}  (CVPR '18)         & 68.2        & 115.0        & 10.6         & 16.0  & 266.1         & 397.5       \\
			ic-CNN \cite{ranjan2018iterative}  (ECCV '18)         & 69.8        & 117.3        & 10.7         & 16.0  & 260.9         & 365.5       \\
			SA-Net \cite{cao2018scale}* (ECCV '18)          & {{67.0}}        & 104.5        &   {{8.4}}         &   {{13.6} }    &  \bf{258.5}        & \bf{334.9}  \\
			IA-DCCN \cite{sindagi2019inverse}* (AVSS '19)          & {{66.9}}        & 108.4        &   {{10.2}}         &   {{16.0} }    &  {264.2}        & {394.4}  \\
			ADCrowdNet \cite{liu2018adcrowdnet}   (CVPR '19)        & {63.2}        & {98.9}        & {\bf{8.2}}         &  15.7      & {266.4}        & 358.0   \\
			RReg \cite{wan2019residual}   (CVPR '19)        & \bf{63.1}        & \bf{96.2}        & {{8.7}}         &  {\bf{13.5}}      & -       & -   \\
			HA-CCN (ours) & \underline{\bf{62.9}}        & \underline{\bf{94.9}}        & \underline{\bf{8.1}}         & \underline{\bf{13.4}}  &   \underline{\bf{256.2}}         & 348.4       \\ \hline
		\end{tabular}
	}
	
\end{table}

\begin{table}[t!]
	\centering
	\caption{Comparison of results on the UCF-QNRF datastet \cite{idrees2018composition}.  Top two methods are highlighted using underline and bold fonts respectively.}
	\label{tab:resultsucf}
	\resizebox{0.7\linewidth}{!}{
		\begin{tabular}{|l|c|c|}
			\hline
			Method & MAE & MSE \\			\hline
			Idrees \etal \cite{idrees2013multi} (CVPR '13) & 315.0 & 508.0 \\
			Zhang \etal \cite{zhang2015cross} (CVPR '15) & 277.0 & 426.0 \\
			CMTL \etal \cite{sindagi2017cnnbased} (AVSS '17)& 252.0 & 514.0 \\
			Switching-CNN \cite{sam2017switching} (CVPR '17) & 228.0 & 445.0 \\
			Idrees \etal \cite{idrees2018composition} (ECCV '18) & {132.0} & {{191.0}} \\
			IA-DCCN \etal \cite{sindagi2019inverse} (AVSS '19) & \bf{125.3} & {\textbf{185.7}} \\
			HA-CCN (ours) & \underline{\textbf{118.1}} & \underline{\bf{180.4 }}\\
			\hline
		\end{tabular}
	}
\end{table}

\begin{figure*}[ht!]	
	\centering	
	\includegraphics[width=0.323\linewidth]{figures/sa/IMG_27}
	\includegraphics[width=0.323\linewidth]{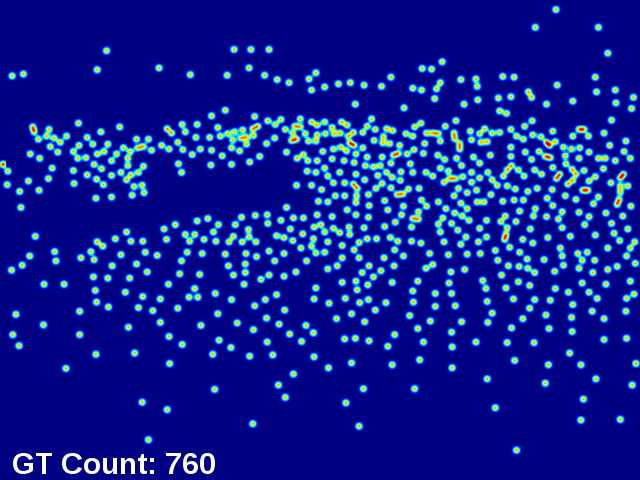}
	\includegraphics[width=0.323\linewidth]{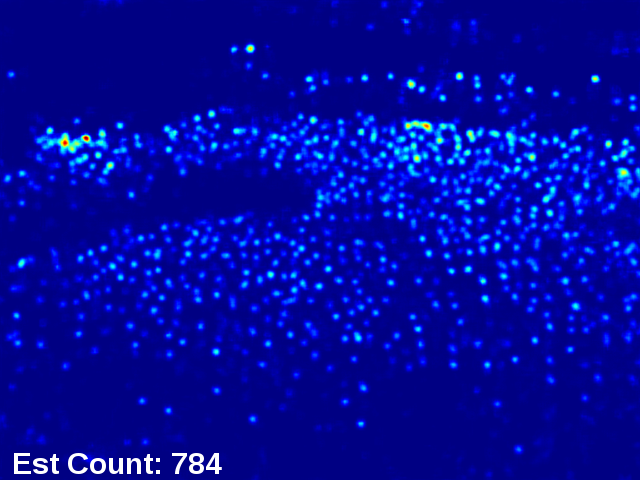}\\
	\vspace{2pt}
	\includegraphics[width=0.323\linewidth]{figures/sa/IMG_30}
	\includegraphics[width=0.323\linewidth]{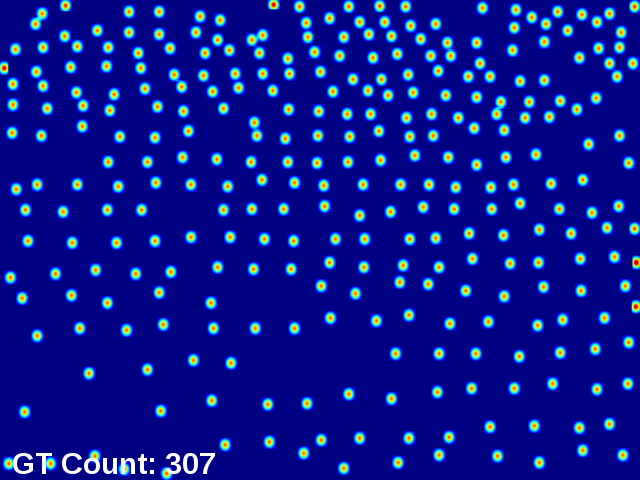}
	\includegraphics[width=0.323\linewidth]{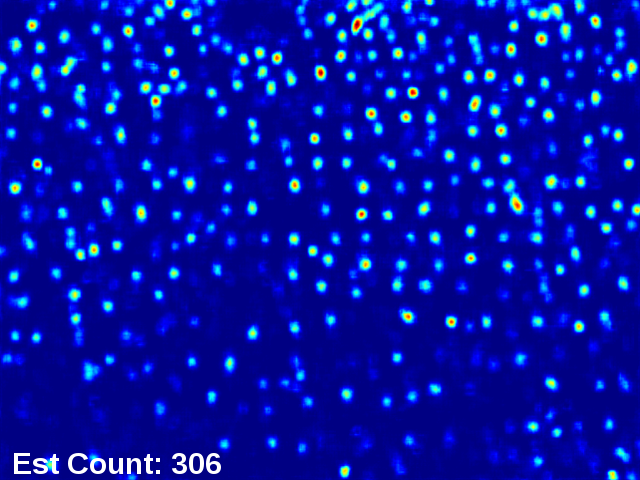}\\
	\vspace{2pt}
    (a)\hspace{160pt}(b)\hspace{160pt}(c)
	\vskip -7pt\caption{Sample results of the proposed method on  ShanghaiTech \cite{zhang2016single}  \textit{(a)} Input. \textit{(b)} Ground truth \textit{(c)} Estimated density map.}
	\label{fig:shtechresults1}	
\end{figure*}
\begin{figure*}[h!]	
	\centering	
	\includegraphics[width=0.323\linewidth]{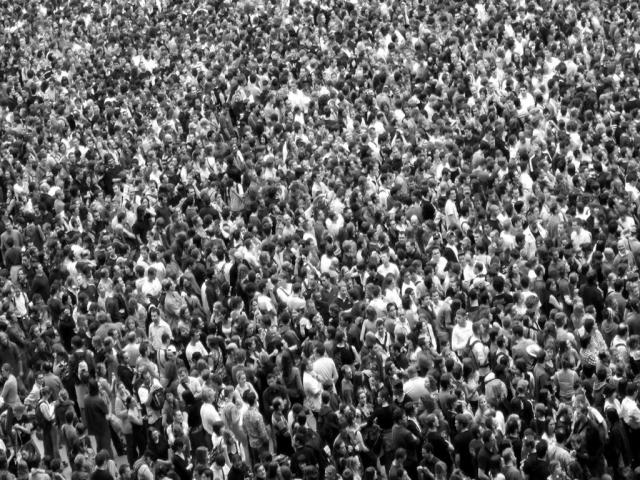}
	\includegraphics[width=0.323\linewidth]{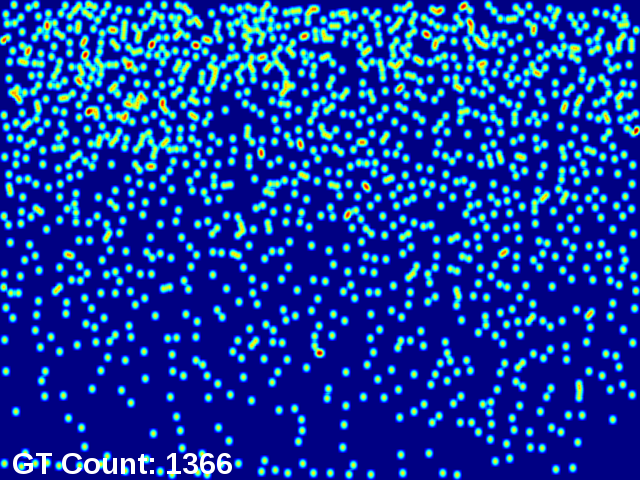}
	\includegraphics[width=0.323\linewidth]{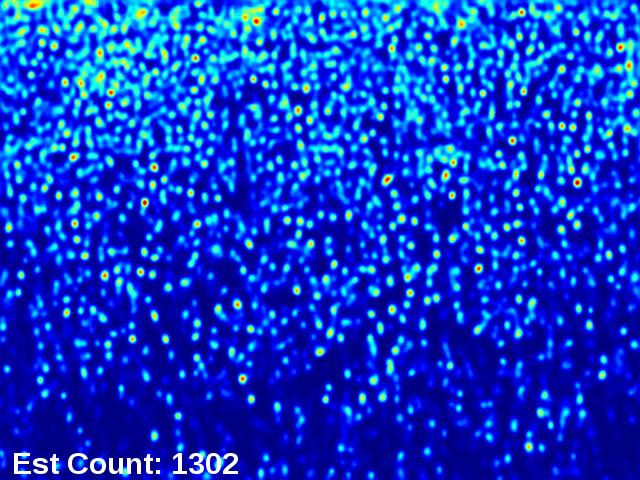}\\
	\vspace{2pt}
	\includegraphics[width=0.323\linewidth]{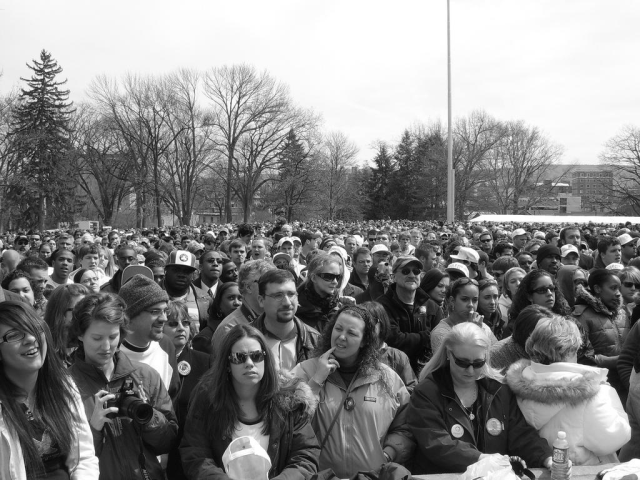}
	\includegraphics[width=0.323\linewidth]{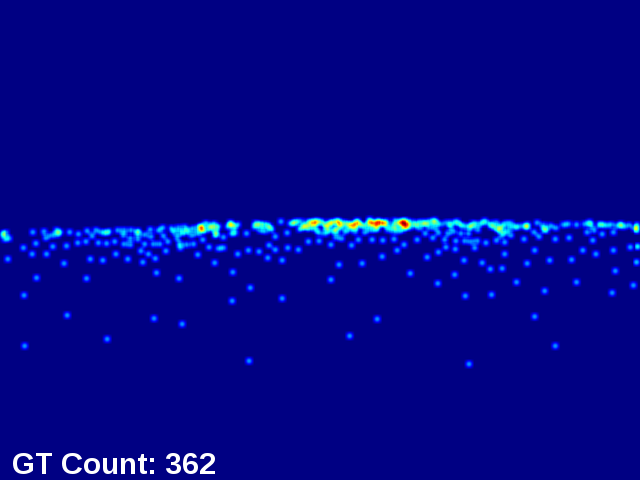}
	\includegraphics[width=0.323\linewidth]{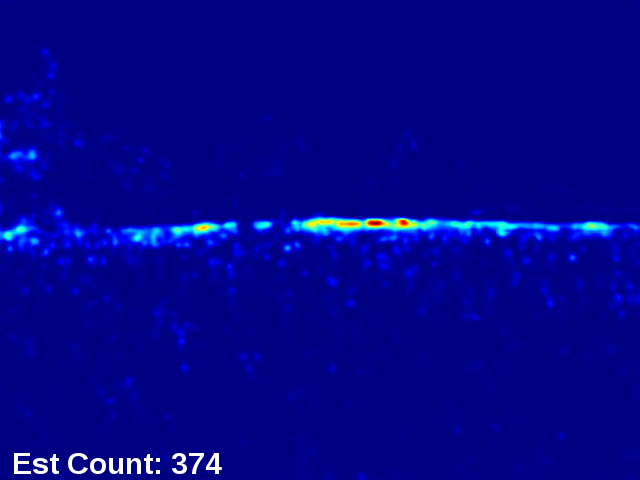}\\
	\vspace{2pt}

	(a)\hspace{160pt}(b)\hspace{160pt}(c)
	\vskip -7pt\caption{Sample results of the proposed method on  \UCF\; \cite{idrees2013multi}.  \textit{(a)} Input. \textit{(b)} Ground truth \textit{(c)} Estimated density map.}
	\label{fig:shtechresults2}	
\end{figure*}

\begin{figure*}[ht!]	
	\centering	
	\includegraphics[width=1\linewidth]{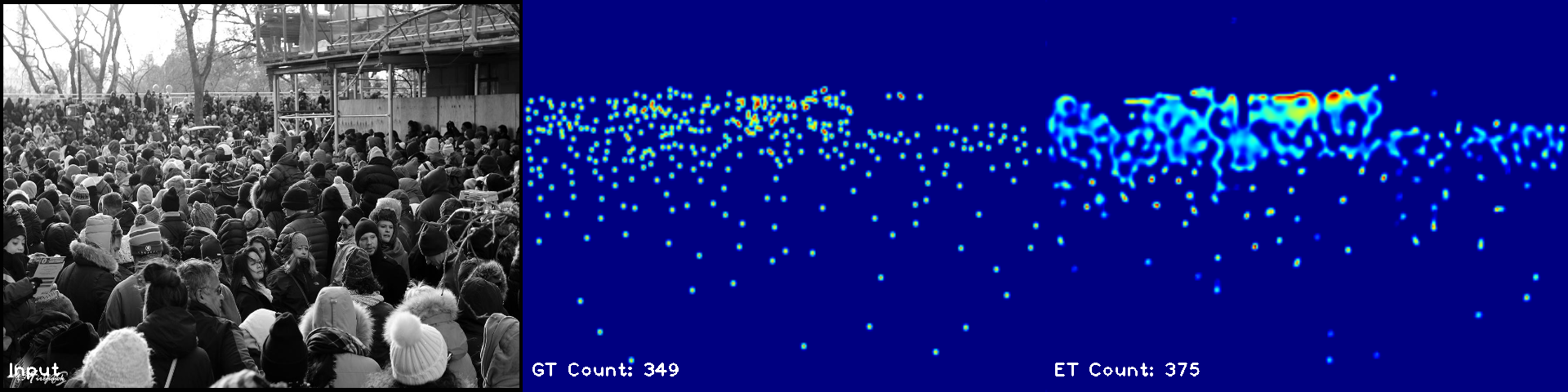}\\
	\vspace{2pt}
	\includegraphics[width=1\linewidth]{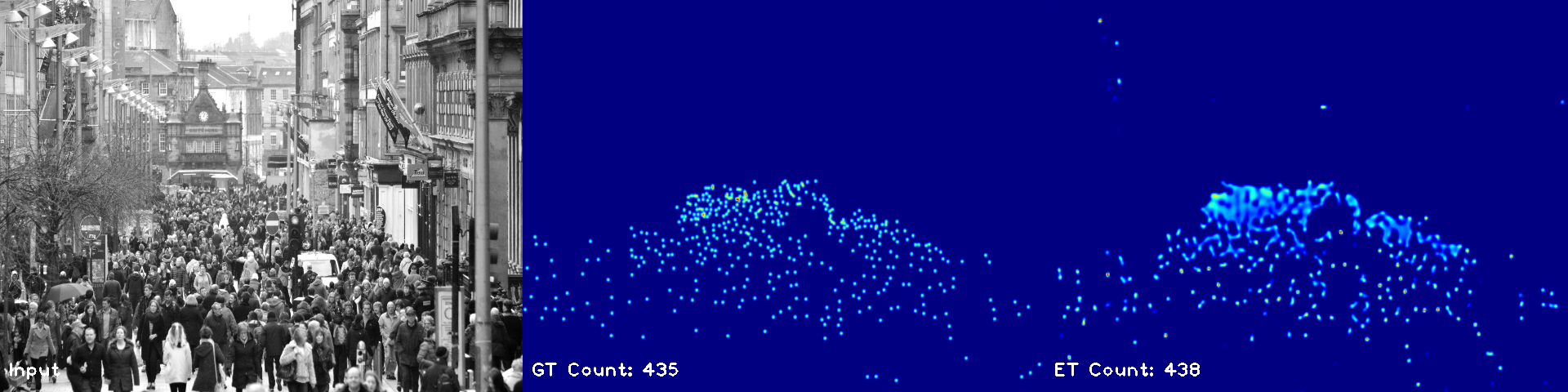}\\
	\vspace{2pt}
	(a)\hspace{160pt}(b)\hspace{160pt}(c)
	\vskip -3pt\caption{Sample results of the proposed method on  UCF-QNRF dataset \cite{idrees2018composition}.  \textit{(a)} Input. \textit{(b)} Ground truth \textit{(c):} Estimated density map.}
	\label{fig:ucfqnrf}	
\end{figure*}
\begin{table*}[ht!]
	\caption{Cross dataset performance. S: Model is trained on target set, NS: Model is trained on source and tested on target set. C: Drop in performance between S and NS.}
	\label{tab:cross_dataset}
	\resizebox{0.95\textwidth}{!}{%
		\begin{tabular}{|l|c|c|c|c|c|}
			\hline
			\multirow{3}{*}{Method}                & \multicolumn{5}{c|}{Target Set}                                                       \\ \cline{2-6} 
			& \multicolumn{2}{c|}{ShanghaiTech B} & \multicolumn{2}{c|}{\UCF}       & WExpo '10 \\ \cline{2-6} 
			& MAE  (S/NS/C)           & MSE   (S/NS/C)            & MAE     (S/NS/C)        & MSE     (S/NS/C)        & MAE  (S/NS/C)          \\ \hline
			MCNN \cite{zhang2016single}            & 26.4/39.6/13.2             & 41.3/102.5/61.2            & 377.6/397.7/20.1          & 509.1/624.1/115.0          & 11.6/25.2/13.6          \\  
			Switch CNN \cite{sam2017switching}  & 21.6/59.4/37.8             & 33.4/130.7/97.3            & 318.1/1117.5/799.4         & 439.2/1315.4/876.2        & 9.4/31.1/21.7          \\  
			D-ConvNet \cite{shi2018crowd_negative} & 18.7/49.1/30.4             & 26.0/99.2/73.2             & 288.4/364.0/75.6          & 404.7/545.8/141.1          & -             \\  
			HA-CCN (ours)                          & {8.1}/{29.1/21.0}    & {13.4}/74.1/{60.1}    & {256.2/339.8/83.6} & {348.4/463.2/114.8} & {8.5/22.0/13.5} \\ \hline
		\end{tabular}%
	}
	\vskip -12pt
\end{table*}

 \subsubsection{Comparison with recent methods}
 
 In this section, we discuss the  results of the proposed method as compared with recent approaches on 3 different datasets: ShanghaiTech \cite{zhang2016single}, \UCF \cite{idrees2013multi} and UCF-QNRF \cite{idrees2018composition}. As discussed earlier, ShanghaiTech has 2 parts with a total of 1198 images.  The UCF\textunderscore CC\textunderscore 50 dataset \cite{idrees2013multi} contains 50 annotated images of different resolutions and aspect ratios. Following the standard protocol discussed in \cite{idrees2013multi}, a 5-fold cross-validation is performed for evaluating the proposed method. UCF-QNRF \cite{idrees2018composition} is a more recent dataset that contains 1,535 high quality images with a total of 1.25 million annotations. The training and test sets consists of 1201 and 334 images respectively.

 Table \ref{tab:shtech} shows the results of the proposed method on the ShanghaiTech and \UCF datasets as compared with several recent approaches: Switching-CNN \cite{sam2017switching},  CP-CNN \cite{sindagi2017generating}, IG-CNN \cite{babu2018divide}, D-ConvNet \cite{shi2018crowd_negative}, Liu \etal \cite{liu2018leveraging}, CSRNet \cite{li2018csrnet}, ic-CNN \cite{ranjan2018iterative}, SA-Net \cite{cao2018scale}, ADCrowdNet \cite{liu2018adcrowdnet} and Residual Regression \cite{wan2019residual}. It can be observed that the proposed method outperforms all existing methods.

 Table \ref{tab:resultsucf} shows the comparison of  results on the recently released large-scale UCF-QNRF \cite{idrees2018composition} dataset. The proposed method is compared against five different approaches: Idrees \etal \cite{idrees2013multi}, MCNN \cite{zhang2016single}, CMTL \cite{sindagi2017cnnbased}, Switching-CNN \cite{sam2017switching} and Idrees \etal \cite{idrees2018composition}. It can be observed that the proposed method is able to achieve state-of-the-art results on this complex dataset. Fig. \ref{fig:shtechresults1}, \ref{fig:shtechresults2} and \ref{fig:ucfqnrf} illustrate the qualitative results for sample images from the ShanghaiTech,  \UCF and UCF-QNFRF datasets respectively.

\subsection{Cross dataset performance}
\label{ssec:cross_dataset}
We compare the generalization abilities of the proposed method with that of  recent methods (MCNN \cite{zhang2016single}, Switching-CNN \cite{sam2017switching}, D-ConvNet \cite{shi2018crowd_negative}) by testing the network (trained on ShanghaiTech A dataset) on target datasets such as ShanghaiTech B, \UCF and WorldExpo '10 \cite{zhang2015cross}. The results are presented in Table \ref{tab:cross_dataset}. Note that the other networks are also trained on ShanghaiTech A dataset. The cross-dataset performance is measured using the overall  count error (MAE/MSE) and the drop in performance. The drop in performance is the difference between the error of the model trained on the target set and that of the model trained on source set, when tested on target set. It can be observed that the proposed method is relatively more robust to change in dataset distribution as compared to the other methods. 

Although the proposed method demonstrates better cross-dataset performance as compared to existing methods, there is considerable gap in the performance as compared to when the network is fully supervised on the target set. We address this issue via the weakly supervised technique described in Section \ref{sec:wsl}.

\subsection{Weakly supervised counting}

In this section, we present the experiment details and results of weak supervision setup.  

\noindent \textbf{Training.} First, a source training set is created that is based on the ShanghaiTech A dataset. The other datasets (ShanghaiTech B, \UCF\; and WorldExpo) are used as the target sets. ShanghaiTech A is chosen for creating the source training set since it contains large variations in density, scale and appearance of people across images. The training set is created by cropping multi-scale patches of size 224$\times$224 from 9 random locations. The multi-scale patch extractions increases diversity of the source dataset in terms of count and field of view. Image-level labels for the source dataset are assigned based on the count in each image in the source set. 

The target training set is created by cropping multi-scale patches from 9 random locations from each image. The image-level labels for the target set are obtained  based on the count in each image. To compensate for the fact that count values from the target set are used to obtain the image-level labels (which is not practically feasible since the target set is not supposed to have point-wise or count annotations), label noise is added for 15\% of the training samples.  That is, we  randomly changed the labels of 15\% of the samples with the neighboring classes. This process of adding label noise simulates human labeling error.

The crowd counting network is first trained on the diverse source dataset using full-supervision by minimizing the loss function  described in  \eqref{eq:lossdensity}, followed by addition of the CAM module. Weights of the counting network are fixed and the CAM module is trained on the diverse source dataset by minimizing the binary cross entropy between image-level labels and aggregated class scores. This is followed by fine-tuning of the CAM module on the target samples using image-level labels. The class-wise maps from the CAM module are used to generate the pseudo ground-truth density maps for the target samples which are then used to fine-tune the counting network.

\noindent \textbf{Discussion.} The results of adapting pre-trained counting model using weak supervision and selective fine-tuning for three target datasets (ShanghaiTech B, \UCF\; and WorldExpo) are shown in Table \ref{tab:wsl}.  For WorldExpo, we average the MAE error over all the five scenes.  For weak supervision, following configurations with three different aggregation functions are evaluated: \\
(1)HA-CCN+W-A: Global Average Pooling (GAP) \\
(2)HA-CCN+W-M: Global Max Pooling (GMP) \\
(3)HA-CCN+W-L: Log-Sum-Exponential (LSE)

It can be observed that the proposed WSL setup results in  significant improvements in the generalization performance of the network. 
Among the three aggregation functions for weakly supervised learning, LSE outperforms the other two functions.  The results obtained using WSL are comparable to many recent fully supervised techniques such as Hydra-CNN \cite{onoro2016towards}, MCNN \cite{zhang2016single}, Walach \etal \cite{walach2016learning}, Switching-CNN \cite{sam2017switching}, thus demonstrating the significance of the proposed weak supervision technique. 
\begin{table}[h!]
	\caption{Results for weakly supervised experiments}
	\label{tab:wsl}
	\resizebox{\textwidth}{!}{%
		\begin{tabular}{|l|c|c|c|c|c|}
			\hline
			\multirow{3}{*}{Method}                & \multicolumn{5}{c|}{Target Set}                                                       \\ \cline{2-6} 
			& \multicolumn{2}{c|}{ShanghaiTech B} & \multicolumn{2}{c|}{\UCF}       & WExpo '10 \\ \cline{2-6} 
			& MAE              & MSE              & MAE            & MSE            & MAE           \\ \hline
			HA-CCN - NS                       & {29.1}    & {74.1}    & {339.8} & {463.2} & {22.0}  \\
			HA-CCN + W-A                           & 23.1             & 50.6             & 320.6       & 430.6       & 17.5 \\
			HA-CCN + W-M                           & 22.5             & 51.2             & 322.2       & 428.1       & 17.7    \\
			HA-CCN + W-L                          & 21.5             & 46.1             & 315.1       & 420.3       & 15.9  \\ \hline
		\end{tabular}%
	}
	\vskip -12pt
\end{table}

\section{Conclusions}
\label{sec:conclusion}

In this work, we presented a crowd counting network that consists of different attention mechanisms at various levels in the network. Specifically, the proposed network involves two sets of attention modules: spatial attention and global attention module.  The spatial attention module incorporates pixel level attention through a way of foreground background segmentation into the features of the earlier layers of the network. The global attention module incorporates channel-wise importance into the network.  Furthermore, we also presented a novel weakly supervised setup to adapt counting models to different datasets using image-level labels.  Extensive  experiments  performed  on  challenging  datasets and  comparison  with  recent  state-of-the-art  approaches demonstrated the significant improvements achieved by the proposed method. 

In the future, we will explore better ways of incorporating features from different layers and extend the current framework to other backbone networks. Additionally, we will explore other forms of weakly supervised and semi-supervised learning approaches to further improve the cross-dataset performance.

\section*{Acknowledgment}
\noindent This work was supported by US Office of Naval Research (ONR) Grant YIP N00014-16-1-3134.

\ifCLASSOPTIONcaptionsoff
  \newpage
\fi



%

	\bibliographystyle{IEEEtran}
	\bibliography{egbib}

\end{document}